\PassOptionsToPackage{table}{xcolor}
\documentclass[11pt, a4paper, logo, copyright, nonumbering]{xiaomi}

\usepackage[authoryear, sort&compress, round]{natbib}
\usepackage{dblfloatfix}
\usepackage{ulem}
\usepackage{caption}
\usepackage{dramatist}
\usepackage{xspace}
\usepackage{pifont}
\usepackage{multirow}
\usepackage{tcolorbox}
\usepackage{xltabular}
\usepackage{longtable}
\usepackage{hyperref}
\usepackage{wrapfig}

\usepackage{amsfonts}
\usepackage{amsmath}
\usepackage{amssymb}
\usepackage{mathtools}
\usepackage{amsthm}
\usepackage{lineno}
\usepackage{adjustbox}

\usepackage[bottom]{footmisc}

\usepackage{CJKutf8}
\usepackage{setspace}
\usepackage{makecell}
\usepackage{graphicx}
\usepackage{subcaption}
\usepackage{multicol}

\usepackage[utf8]{inputenc}
\usepackage[T1]{fontenc}
\usepackage{cleveref}
\usepackage{url}
\usepackage{booktabs}
\usepackage{nicefrac}
\usepackage{microtype}
\usepackage{xcolor}
\usepackage{colortbl}

\definecolor{BrickRed}{rgb}{.72,0,0}
\definecolor{darkgreen}{rgb}{0.0, 0.5, 0.0}
\definecolor{ForestGreen}{RGB}{34,139,34}
\definecolor{LakeBlue}{RGB}{0,61,153}
\definecolor{MiOrange}{RGB}{255,225,204}
\definecolor{Hex}{RGB}{225,213,231}

\hypersetup{
    colorlinks=true,
    allcolors=LakeBlue,
    pdftitle={Beyond Binary: Reframing GUI Critique as Continuous Semantic Alignment},
    pdfauthor={Yuchen Sun, Pei Fu, Shaojie Zhang, Xiuwen Xi, Ruoceng Zhang, Zhenbo Luo, Jian Luan, Heng Qu, Chongyang Zhang}
}

\theoremstyle{plain}

\theoremstyle{definition}

\theoremstyle{remark}

\definecolor{rowgray}{gray}{0.92}

\newcommand{\gain}[1]{%
  \hspace{1pt}%
  {\fontsize{7pt}{1em}\selectfont\textbf{+}#1}%
}

\newcommand{\steady}{%
  \hspace{1pt}%
  {\fontsize{7pt}{1em}\selectfont\textbf{--}}%
}

\newcommand{\cmark}{\ding{51}}
\newcommand{\xmark}{\ding{55}}

\renewcommand{\thefootnote}{\fnsymbol{footnote}}

\title{\centering Beyond Binary: Reframing GUI Critique as Continuous Semantic Alignment\\[8pt]{\normalfont\normalsize\itshape Accepted to AAAI 2027}}

\author[1]{Yuchen Sun\thanks{Work done during an internship at Xiaomi Inc.}}
\author[2]{Pei Fu}
\author[2]{Shaojie Zhang}
\author[2]{Xiuwen Xi}
\author[2]{Ruoceng Zhang}
\author[2]{Zhenbo Luo\thanks{Corresponding authors.}}
\author[2]{Jian Luan}
\author[2]{Heng Qu}
\author[1]{Chongyang Zhang\textsuperscript{$\dagger$}}
\affil[1]{Shanghai Jiao Tong University}
\affil[2]{Xiaomi Inc.}

\begin{document}

\begin{abstract}
Test-Time Scaling (TTS), which samples multiple candidate actions and ranks them via a Critic Model (which scores how well each candidate UI action advances the user's task), has emerged as a promising paradigm for generalist GUI agents. Its efficacy thus hinges on the critic's fine-grained ranking ability. However, existing GUI critic models uniformly adopt binary classification. Our motivational analysis of these models exposes a severe entanglement: scores for valid actions and plausible-but-invalid distractors become indistinguishable. We attribute this failure to two structural defects: \textit{Affordance Collapse}---the hierarchical affordance space is compressed into 0/1 labels; and \textit{Noise Sensitivity}---binary objectives overfit to noisy decision boundaries. To resolve this, we introduce \textbf{BBCritic} (\underline{B}eyond-\underline{B}inary \underline{C}ritic), a reformulation grounded in the Functional Equivalence Hypothesis. Through two-stage contrastive learning, BBCritic aligns instructions and actions in a shared Affordance Space, recovering the hierarchical structure that binary supervision flattens. We also present \textbf{BBBench} (\underline{B}eyond-\underline{B}inary \underline{B}ench), the first GUI critic benchmark that pairs a dense action space with a hierarchical four-level taxonomy, enabling fine-grained ranking evaluation. Experimental results show that BBCritic-3B, trained without any extra annotation, outperforms 7B-parameter SOTA binary models. It demonstrates strong zero-shot transferability across platforms and tasks, supporting our methodological view: GUI critique is fundamentally a metric-learning problem, not a classification one.
\end{abstract}

\maketitle
\makeatletter
\@thanks
\let\@thanks\@empty
\makeatother

\renewcommand{\thefootnote}{\arabic{footnote}}
\setcounter{footnote}{0}

\section{Introduction}
\begin{figure}[ht]
  \vspace{-5pt}
  \centering
    \includegraphics[width=\textwidth]{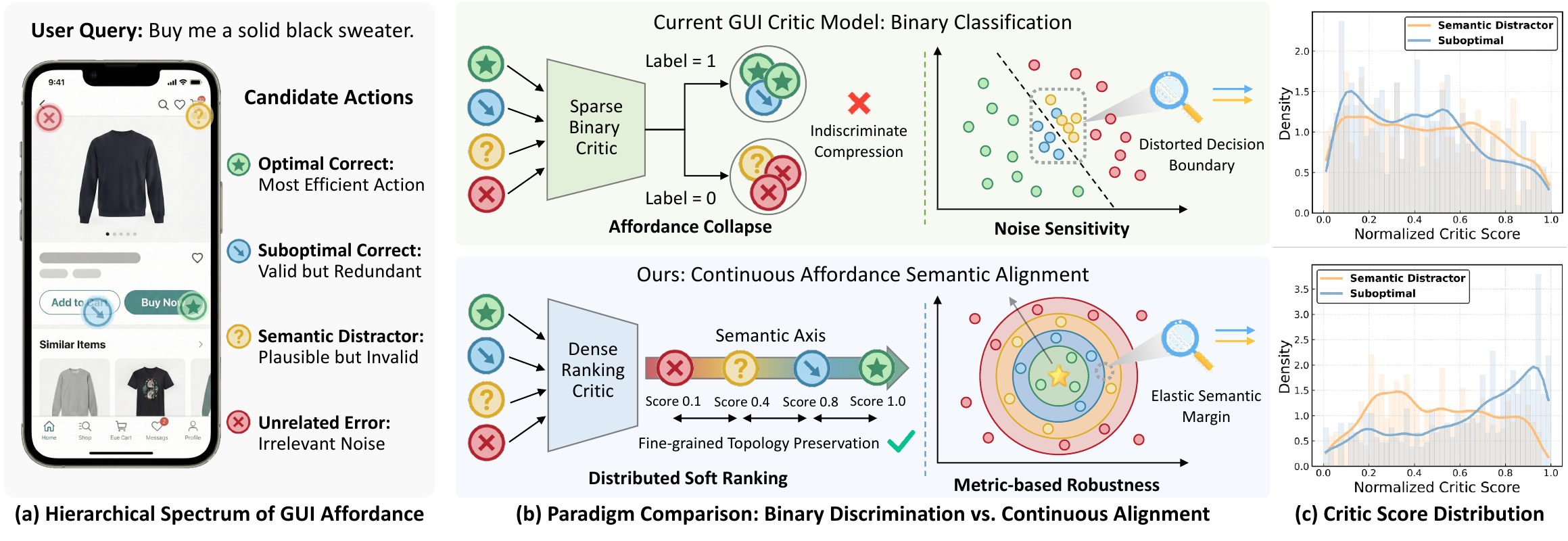}
    \caption{
    Motivational analysis and empirical observation. (a) Motivational Example. Instead of binary labels, human intuition evaluates action validity as a continuous spectrum based on \textbf{functional alignment}. (b) GUI Critic Paradigm Comparison. Current methods relying on binary classification lead to \textbf{Affordance Collapse and Noise Sensitivity}. In contrast, our approach achieves robust ranking via \textbf{continuous affordance alignment}. (c) Score Distribution. Empirical results verify that our method preserves the discriminative margin between Suboptimal and Semantic Distractor, resolving the confusion seen in baselines.
    }
    \label{fig:motivation}
    \vspace{-15pt}
\end{figure}
Generalist GUI agents are transitioning from behavior cloning to verifiable reasoning~\cite{luo2025vimo, yang2025omniactor}. While large-scale pre-training confers foundational capabilities~\cite{sun2024hunyuan, team2023gemini}, randomness in generation often creates error cascades in long-horizon GUI tasks~\cite{chen2025d}. To mitigate this, Test-Time Scaling (TTS)~\cite{yang2025gta1} has emerged as a key paradigm, employing an independent Critic Model to rank candidate actions by how well each advances the user's task. Consequently, the Critic's ranking ability becomes the bottleneck for TTS reliability.

Existing approaches uniformly frame this critique task as binary classification~\cite{wanyan2025look, wang2026gaia, xiao2025ui}. However, the validity of a GUI action is not binary, but lies on a hierarchical spectrum. As shown in Figure~\ref{fig:motivation}(a), for the task ``Buy a solid black sweater,'' clicking ``Buy Now'' is optimal, while ``Add to Cart'' is valid but redundant. Meanwhile, clicking the cart icon constitutes a semantic distractor—logically related but incorrect—yet distinct from the completely irrelevant ``Back'' action. Distinguishing these gradations requires \textbf{GUI Affordance Modeling}~\cite{norman1999affordance}---evaluating the functional alignment between action and intent, not just visual matching. The binary formulation effectively compresses this continuous spectrum into a rigid 0/1 boundary. Empirically, our motivational analysis reveals that this binary paradigm leads to a breakdown in affordance modeling, a failure we attribute to two structural defects: \textbf{1. Affordance Collapse:} Binary supervision compresses the hierarchical affordance space into 0/1 labels, blurring both the ordering within each polarity and the boundary between them. The critic thus fails to distinguish exploratory yet valid attempts from plausible errors. \textbf{2. Noise Sensitivity:} The boundary between ``correct'' and ``wrong'' is often inherently fuzzy. Binary classifiers, with their absolute penalties, cannot accommodate this ambiguity and easily overfit to annotation noise.

To resolve these defects, we introduce \textbf{BBCritic} (\underline{B}eyond-\underline{B}inary \underline{C}ritic), a reformulation grounded in a \textit{Functional Equivalence Hypothesis}. We posit that the user instruction and the optimal action are two expressions of the same underlying intent—one linguistic and one operational—sharing the same functional affordance. Guided by this insight, BBCritic reframes GUI critique from rigid binary discrimination to continuous semantic alignment via contrastive learning. By optimizing instruction-action similarity in a shared Affordance Space, our framework recovers the action hierarchy: valid solutions cluster together while unrelated errors are pushed apart. In a weakly supervised setting without extra annotations, BBCritic-3B significantly outperforms 7B-parameter SOTA binary models across diverse benchmarks. Notably, this formulation also exhibits strong robustness to label noise and cross-platform generalization.

Existing GUI critic benchmarks~\cite{wanyan2025look, wu2025oracle} typically annotate only a single action per page with a binary label, leaving them unable to evaluate the fine-grained ranking ability essential for TTS. To bridge this gap, we introduce \textbf{BBBench} (\underline{B}eyond-\underline{B}inary \underline{B}ench), a GUI critic benchmark with two core properties: \textbf{(i)} a dense action space that exhaustively covers all feasible UI actions on a page ($\sim$30 candidates on average), and \textbf{(ii)} a hierarchical four-level taxonomy (Optimal, Suboptimal, Semantic Distractor, Unrelated Error) that maps the continuous affordance spectrum into evaluable tiers. With 18,192 human-verified samples, BBBench is the first benchmark to test whether GUI critic scores consistently reflect functional quality across the full action space.

Our main contributions are summarized as follows.

\textbf{(1) Reformulation of GUI Critique: } We are the first to model GUI critique from an affordance perspective. Built on the Functional Equivalence Hypothesis, our contrastive framework reframes critique as continuous semantic alignment, recovering the action hierarchy and resolving the affordance collapse of binary baselines.
  
\textbf{(2) The BBBench Benchmark: }We establish BBBench, the first GUI critic benchmark that pairs a dense action space with a hierarchical four-level functional taxonomy. This design enables direct evaluation of whether critic scores monotonically reflect action quality---a property essential for ranking-based GUI agents.

\textbf{(3) Performance Verification: }Experiments show that BBCritic achieves consistent improvements and generalization across platforms and tasks—using half the parameters of baselines (3B vs. 7B) and zero extra annotations.

\section{Related Work}

\textbf{GUI Agents.} GUI agents have advanced rapidly~\cite{chen2025ui, liu2025scalecua}, driven by vision-language generalists capable of interacting with diverse interfaces~\cite{seed2025seed1,guan2025kg}. Yet their reliability in long-horizon tasks remains constrained by error cascading~\cite{chen2025d, chai2025a3}. To mitigate this, research has split into two directions: capability internalization, which embeds reasoning via post-training~\cite{shi2025mobilegui, liu2025infigui}, and Test-Time Scaling, which trades inference compute for decision quality via search or sampling~\cite{wu2025gui,wang2024mobile}. Our work aligns with the latter. However, the efficacy of TTS is strictly bounded by the critic's discriminative capacity, motivating intensive research on robust critic models.

\textbf{GUI Critic Models.}
Critic Models play a central role in aligning model outputs with human intent~\cite{mcaleese2024llm, wanyin2025self}, and contrastive learning has proven effective for mapping abstract representations to human preferences~\cite{chen2024improving, lin2025hif}. In the GUI domain, however, methodologies remain underdeveloped. While UITARS2~\cite{wang2025ui} predicts outcome-level supervision, other methods focus on fine-grained step-level verification. These approaches employ diverse data strategies, ranging from human-annotated feedback (GUI-Shepherd~\cite{chen2025gui}) to scalable synthetic generation (UI-Genie~\cite{xiao2025ui}, GAIA~\cite{wang2026gaia}, OS-Oracle~\cite{wu2025oracle}). Yet, regardless of granularity or data source, existing methods predominantly frame evaluation as a binary classification task, which fails to capture the hierarchical nature of GUI affordances. We address this gap with a contrastive framework that recovers the continuous functional topology of the action space under weak supervision.

\textbf{GUI Critic Benchmark.}
Existing protocols for GUI agents primarily evaluate execution success in offline~\cite{yang2025fingertip, yang2025probench} or online environments~\cite{xie2024osworld, zhou2023webarena}. As GUI critic models emerged, dedicated benchmarks followed, such as GUI-Critic-Test~\cite{wanyan2025look}, OS-Oracle~\cite{wu2025oracle}, and UI-Genie~\cite{xiao2025ui}. However, these benchmarks typically annotate only a single action per page with a binary label, failing to test whether a critic truly understands GUI action semantics---the essence of critic capability. To bridge this gap, we introduce BBBench, a GUI critic benchmark that pairs a dense action space with a hierarchical four-level functional taxonomy. Unlike binary predecessors, BBBench enables direct evaluation of whether critic scores monotonically reflect functional quality across the full action space.
\section{Problem Formulation \& Motivation}

\subsection{Task Formulation}
We formulate the Test-Time Scaling process for GUI agents as a Partially Observable Markov Decision Process (POMDP), defined by the tuple ( $\mathcal{I}, \mathcal{S}, \mathcal{A}, \mathcal{O}, \mathcal{T}$ ). Here, $\mathcal{I}$ represents the high-level user instruction, $\mathcal{S}$ denotes the latent state space, $\mathcal{A}$ is the action space, $\mathcal{O}$ is the observation space, and $\mathcal{T}: \mathcal{S} \times \mathcal{A} \rightarrow \mathcal{S}$ is the transition function.

At each time step $t$, the Policy Model $\mathcal{M}_{\text {policy }}$ samples a set of $N$ candidate actions $\left\{a_t^i\right\}_{i=1}^N \subset \mathcal{A}$ based on the instruction $\mathcal{I}$, the interaction history $h_t=\left\{\left(o_j, a_j\right)\right\}_{j=0}^{t-1}$, and the current observation $o_t$. Subsequently, the Critic Model $\mathcal{M}_{\text {critic }}$ evaluates and ranks these candidates to select the optimal action $a_t^*$ :
$$
\begin{gathered}
\left\{a_t^i\right\}_{i=1}^N \sim \mathcal{M}_{\text {policy }}\left(\cdot \mid \mathcal{I}, h_t, o_t\right) \\
s_t^i=\mathcal{M}_{\text {critic }}\left(\mathcal{I}, h_t, o_t, a_t^i\right) \\
a_t^*=\underset{a_t^i}{\arg \max }\ s_t^i
\end{gathered}
$$
where $s_t^i \in \mathbb{R}$ is a continuous critic score quantifying the semantic alignment between candidates and user instruction.

\subsection{The Binary Bottleneck: Empirical Analysis} \label{motivation}

Within this formulation, the efficacy of TTS is strictly bounded by the Critic Model's ability to rank candidate actions. Existing methods~\cite{wang2026gaia} adapt binary classifiers by treating the logit of the ``Correct'' token as a proxy scalar score. To test whether this binary proxy captures the underlying semantic structure, we visualize the critic score distribution on BBBench (Sec.~\ref{sec:bbbench}), using sigmoid-normalized logits for binary baselines and cosine similarity for BBCritic. We focus on the most ambiguous decision boundary: \textit{Suboptimal} actions (valid but inefficient) versus \textit{Semantic Distractors} (plausible but functionally invalid).

The results in Figure~\ref{fig:motivation}(c) reveal a stark gap between binary classification accuracy and genuine semantic modeling. We observe a severe \textit{distributional entanglement} where the critic scores for suboptimal actions and semantic distractors are indistinguishable. This overlap exposes a structural weakness: binary objectives only enforce a coarse Optimal-vs-Unrelated split, lacking the fine-grained \textit{GUI Affordance Modeling} required to map action quality to a continuous space. Compounding the issue, current benchmarks share the same blind spot---they too evaluate only binary correctness and thus cannot expose this granularity gap.

\begin{figure}[ht]
  \centering
    \includegraphics[width=\textwidth]{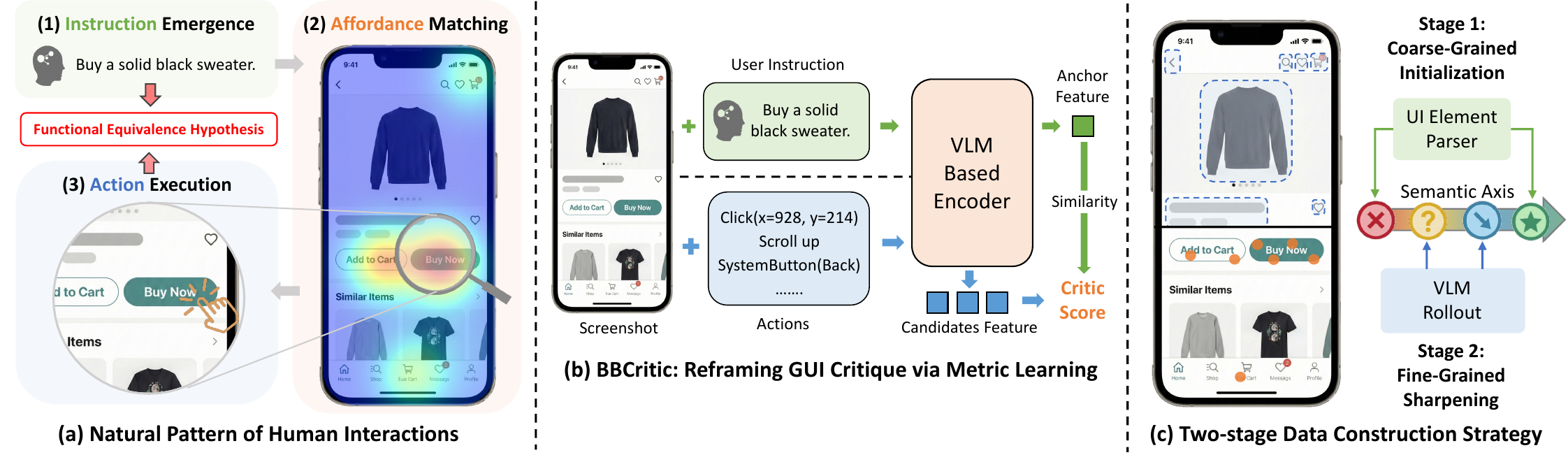}
    \caption{
    Overview of the Contrastive Semantic Alignment Framework. (a) Natural Pattern of Human Interactions: Instruction and Action are framed as two forms of the same target Affordance via the Functional Equivalence Hypothesis. (b) BBCritic -- Reframing GUI Critique via Metric Learning: A shared VLM encoder maps inputs into a shared Affordance Space to quantify their alignment. (c) Two-stage Data Construction Strategy: A coarse-to-fine curriculum using layout parsing for initialization and VLM rollouts for boundary sharpening.
    }
    \label{fig:method}
    \vspace{-12pt}
\end{figure}
\section{Method}

Motivated by the empirical limitations observed in Sec.~\ref{motivation}, we analyze the structural incompatibility between rigid binary objectives and the hierarchical nature of GUI Affordance. We then introduce our contrastive framework, grounded in the \textit{Functional Equivalence Hypothesis}, designed to recover the fine-grained structure of the Affordance Space.

\subsection{Theoretical Analysis: The Binary Mismatch} \label{sec:binary_mismatch}

As shown in Figure~\ref{fig:motivation}, GUI action validity forms a hierarchical spectrum, yet existing Critic Models compress it into a rigid binary partition. We identify the structural root of this failure as the \textit{Gradient Independence} inherent in the Binary Cross-Entropy (BCE) loss. For a negative candidate $a^-$, the BCE loss $\mathcal{L}_{\mathrm{BCE}} = -\log(1-\sigma(s^-))$ yields the gradient $\frac{\partial \mathcal{L}_{\mathrm{BCE}}}{\partial s^{-}}=\sigma(s^{-})$. This magnitude depends solely on the sample's own score, decoupled from the positive anchor and other candidates. BCE ranks well under clean labels~\citep{agarwal2014surrogate}; in the noisy GUI critic setting, this isolation causes two failures:

\textbf{Affordance Collapse.} GUI errors are hierarchical: for a ``Buy'' intent, clicking ``Cart'' is a semantic distractor, whereas ``Back'' is irrelevant. While a critic should preserve this topological distance, BCE is topology-agnostic due to gradient independence. It suppresses all negatives equally and pulls all positives equally, compressing the diverse functional hierarchy into a flat binary partition. Consequently, the model loses both the intra-polarity hierarchy (Optimal vs. Suboptimal, Distractor vs. Unrelated) and the cross-polarity boundary between Suboptimal and Distractor.

\textbf{Noise Sensitivity.} Unlike rigid object classification, GUI decision boundaries harbor an intrinsic semantic ``gray zone.'' For instance, when looking for product specifications, clicking ``User Reviews'' instead of the ``Specifications'' tab is semantically relevant yet functionally indirect. BCE, however, cannot accommodate this ambiguity. It forces such plausible exploration into hard negative labels with saturated gradient penalties ($\sigma(s^-) \approx 1$). By treating exploratory yet valid attempts identically to genuine errors, the objective compels the model to overfit to noisy labels rather than learning a reliable scoring function.

\subsection{BBCritic: GUI Critique Beyond Binary Labels} \label{hypos}

To address the limitations of binary classification, we propose a reformulation motivated by \textit{Norman's Action Cycle}~\cite{pea1987user} (Figure~\ref{fig:method}). This framework characterizes human interaction as a logical progression: users formulate a high-level \textbf{Instruction}, search for the \textbf{Affordance} that realizes it, and trigger the \textbf{Action}.

This sequential continuity implies that the intent and the execution are linked by the target affordance. Based on this, we formulate the \textit{Functional Equivalence Hypothesis} for GUI critique: The linguistic instruction $\mathcal{I}$ and the optimal action $a^*$ are two forms of the same underlying GUI Affordance—one as the \textbf{textual description}, the other as the \textbf{behavioral execution}. Accordingly, we reframe the critic's objective from binary classification to measuring continuous semantic alignment. Specifically, we project both modalities into a shared embedding space, which we term the \textit{Affordance Space}.

\textbf{Unified Latent Projection.}
To realize this hypothesis, we adopt a contrastive framework inspired by Universal Multimodal Retrieval~\cite{zhang2024gme, jiang2024vlm2vec} (see Appendix~\ref{app:umr} for background). We repurpose a pre-trained Vision-Language Model from a generative predictor to a dense encoder that maps both instruction and action into this shared Affordance Space.

Formally, we employ a Siamese architecture with a shared VLM backbone. Following GME~\cite{zhang2024gme}, we extract the final hidden state of the last valid token as the feature representation. The projection is defined as follows:

\textit{Intent Anchor ($v_{i}$):} We feed the User Instruction $\mathcal{I}$ concatenated with the current Screenshot $o_t$ into the encoder. The resulting embedding $v_i$ serves as the holistic representation of the \textbf{desired affordance}.

\textit{Action Candidate ($v_{a}^i$):} We feed the Candidate Action $a^i$ integrated with the Screenshot $o_t$ (with Set-of-Mark visual prompting; see Appendix~\ref{app:som}) into the encoder. The resulting embedding $v_a^i$ serves as the contextualized representation of the \textbf{available affordance}.

\textbf{Continuous Alignment Metric.}
Finally, the critic score for candidate $i$ is the cosine similarity $s^i = \cos(v_i, v_a^i)$ between the intent and action embeddings, later combined with a temperature $\tau$ inside the InfoNCE objective below. Unlike binary logits, this continuous metric preserves the ranking structure of the action space, providing fine-grained signals essential for effective test-time scaling.

\subsection{Topology-Aware Optimization}
To recover the fine-grained hierarchy within the Affordance Space, we optimize using the InfoNCE loss. Unlike binary objectives that evaluate candidates in isolation, InfoNCE introduces a distributional dependency across the dense action space. Formally, the loss is defined as $\mathcal{L} = -\log \frac{\exp(s^+/\tau)}{\exp(s^+/\tau) + \sum_{k} \exp(s^{k-}/\tau)}$. The gradient for a negative candidate $a^{k-}$ is derived as $\frac{\partial \mathcal{L}}{\partial s^{k-}} = \frac{1}{\tau} P(a^{k-})$, where $P(a^{k-})$ is the softmax-normalized probability of treating $a^{k-}$ as the positive within the batch. This gradient mechanism is well-suited for GUI critique:

\textbf{Mining Hard Negatives in Dense Action Spaces.} GUI pages are characterized by a severe imbalance: a single optimal action exists amidst dozens of negative elements. Binary loss tends to be overwhelmed by easy negatives (e.g., irrelevant background), leading to gradient vanishing on subtle errors. In contrast, the InfoNCE gradient is proportional to the relative confusion $P(a^{k-})$. This automatically directs optimization focus toward semantic hard negatives—actions that are functionally similar to the intent but incorrect. This ``self-paced'' mining forces the model to sculpt a precise decision boundary, resolving the \textit{Affordance Collapse}.

\textbf{Robustness to Ambiguity via Gradient Coupling.} As analyzed in Sec.~\ref{sec:binary_mismatch}, GUI operations inherently contain semantic ``gray zones'' that confuse absolute binary classifiers. InfoNCE mitigates this by enforcing relative ranking ($s^+ > s^-$) rather than absolute decisions. The penalty for a negative sample is coupled with the confidence in the positive anchor. This implies that as long as the optimal action is ranked highest, the model is not forced to crush the score of a Suboptimal action to zero. This relative constraint makes the critic robust to label noise, preventing overfitting to noisy annotations.

We adopt InfoNCE over alternative ranking objectives (pairwise, listwise, ordinal); see Appendix~\ref{app:infonce_vs_pairwise} for justification.

\subsection{Training Data Construction Strategy} \label{twostage}

We design a two-stage training curriculum that moves from coarse coverage to fine-grained semantic discrimination.

\textbf{Stage 1: Coarse-Grained Topology Initialization.} The primary objective of this stage is to establish a comprehensive understanding of the GUI interaction space. A robust critic must first distinguish the optimal action from the vast background of irrelevant elements. To achieve this, we utilize Ground Truth (GT) actions as positive anchors. To exhaustively cover the feasible action space, we parse the full UI layout via OmniParserV2~\cite{yu2025omniparser} and treat all non-target interactive elements as dense negatives. Non-click actions (e.g., scroll, type) are added through rule-based generation. This coarse-grained stage establishes the model's basic ability to isolate optimal actions from the dense candidate pool.

\textbf{Stage 2: Fine-Grained Boundary Sharpening.} Building upon this coarse topology, the second stage targets the cross-polarity boundary (e.g., suboptimal-vs-distractor)---the most ambiguous face of Affordance Collapse. We employ a generic Vision-Language Model to perform Heuristic Rollouts on the training set. Sampled actions that differ from the recorded ground-truth action at the same step are harvested as ``Semantic Hard Negatives.'' These samples represent plausible but incorrect attempts. Incorporating these samples sharpens the decision boundaries in the Affordance Space.

\textbf{Weak Supervision Protocol.} To validate the robustness of our framework, this entire pipeline follows a strict weak-supervision setting. We utilize only existing navigation trajectories and self-generated rollouts, introducing zero additional annotation for training.

\begin{table}[t]
  \centering
  \caption{Comparison with existing GUI Critic benchmarks. BBBench uniquely combines a dense per-page action space with a 4-level semantic hierarchy.}
  \label{tab:dataset-comparison}
  
  \small
  \setlength{\tabcolsep}{6pt} 
  
  \begin{tabular}{ccccc}
    \toprule
    \raisebox{0.5\baselineskip}{\textbf{Dataset}} &
    \textbf{\shortstack{Sample\\Num.}} & 
    \textbf{\shortstack{Avg.\\Action}} & 
    \textbf{\shortstack{Critic\\Level}} & 
    \textbf{\shortstack{Multi.\\Pos.}} \\
    \midrule
    
    GUI-Critic-Test & 1,188 & 1.00 & 2 & No \\
    UI-Genie        & 2,100 & 2.00 & 2 & No \\
    OS-Oracle       & 738   & 1.00 & 2 & No \\
    
    \rowcolor{rowgray}
    \textbf{BBBench}   & \textbf{18,192} & \textbf{30.78} & \textbf{4} & \textbf{Yes} \\
    \bottomrule
  \end{tabular}
  \vspace{-10pt}
\end{table}
\section{BBBench: A Fine-Grained GUI Critique Benchmark} \label{sec:bbbench}

\begin{table}[t]
  \centering
  \caption{Performance comparison on BBBench with Ranking (NDCG) and Pairwise Preference (PPA) metrics. Best results within each model category (General VLMs and Specialized GUI Policy Models) are highlighted in \textbf{bold}.}
  \label{tab:critic-performance}
  
  \small
  \setlength{\tabcolsep}{4pt}

  \resizebox{\textwidth}{!}{%
  \begin{tabular}{lcccccc}
    \toprule
    \textbf{Model} & 
    \textbf{NDCG@8} & \textbf{NDCG@16} & \textbf{NDCG@All} & 
    $\textbf{PPA}_{\textbf{opt-sub}}$ & $\textbf{PPA}_{\textbf{sub-dis}}$ & $\textbf{PPA}_{\textbf{dis-unr}}$ \\
    \midrule
    
    \multicolumn{7}{c}{\textit{General VLMs}} \\
    \midrule 
    Gemini 3 Flash~\cite{google_gemini_3_2026}      & 55.13 & 55.32 & 55.50 & 66.60 & 50.69 & 66.28 \\
    Gemini 3 Pro~\cite{google_gemini_3_2026}        & 70.92 & 71.86 & 72.62 & 80.66 & \textbf{62.47} & 71.35 \\
    Claude Sonnet 4.0~\cite{anthropic_claude_4_2026}   & 64.52 & 69.86 & 76.75 & \textbf{86.90} & 47.17 & 66.13 \\
    GPT-4o~\cite{openai_gpt5_2026}              & 74.85 & 76.04 & 80.68 & 80.26 & 47.96 & 69.56 \\
    GPT-5~\cite{openai_gpt5_2026}               & \textbf{78.25} & \textbf{80.49} & \textbf{84.52} & 77.38 & 59.81 & \textbf{74.96} \\
    Qwen2.5-VL-7B~\cite{bai2025qwen2}       & 67.04 & 71.53 & 77.78 & 74.32 & 46.90 & 57.62 \\
    Qwen3-VL-8B~\cite{yang2025qwen3}         & 63.00 & 69.65 & 75.23 & 75.23 & 52.31 & 74.43 \\
    
    \midrule
    \multicolumn{7}{c}{\textit{Specialized GUI Policy Models}} \\
    \midrule
    UITARS-1.5-7B~\cite{qin2025ui}  & 38.35 & 47.70 & 56.78 & 71.36 & 51.98 & 57.19 \\
    UI-Venus~\cite{gu2025ui}       & 36.12 & 45.55 & 57.03 & 60.39 & 46.80 & 56.89 \\
    MAI-UI~\cite{zhou2025mai}         & 43.11 & 51.75 & 61.83 & 66.62 & 40.23 & 61.40 \\
    GUI-Critic-R1~\cite{wanyan2025look}  & 6.4   & 17.62 & 38.51 & 26.53 & 47.56 & 37.03 \\
    GAIA~\cite{wang2026gaia}           & 46.20 & 55.31 & 63.48 & 77.17 & 45.65 & 55.32 \\
    
    \rowcolor{rowgray}
    \textbf{BBCritic-3B} & 70.48 & 75.07 & 80.56 & \textbf{80.99} & 51.20 & 66.10 \\
    \rowcolor{rowgray}
    \textbf{BBCritic-7B} & \textbf{75.09} & \textbf{78.67} & \textbf{83.19} & 78.72 & \textbf{52.67} & \textbf{74.10} \\
    
    \bottomrule
  \end{tabular}}
  \vspace{-12pt}
\end{table}

As discussed in Sec.~\ref{motivation}, current benchmarks share the same binary blind spot as the methods they evaluate, leaving critics' true semantic understanding untested. To fill this gap, we introduce \textbf{BBBench}, a GUI critic benchmark that pairs a dense action space with a hierarchical taxonomy.
\subsection{Data Construction \& Hierarchical Taxonomy}

Two design conditions guide our construction of BBBench: \textbf{(i)} a \textit{dense action space} covering diverse candidates, and \textbf{(ii)} a \textit{hierarchical annotation system} reflecting varying functional validity. We address both through the pipeline below.

\textbf{Dense Action Space.} We build upon MobiBench~\cite{im2025modular}, which provides multiple valid actions per state. To transition from sparse ground truth to a dense evaluation space, we employ a similar candidate expansion strategy introduced in Sec.~\ref{twostage}. Specifically, we combine the XML view hierarchy with OmniParserV2~\cite{yu2025omniparser} detections to extract all interactable elements (e.g., buttons, links, input fields) as negative candidates, ensuring global spatial coverage. Additionally, to simulate the deceptive nature of real-world agent errors, we incorporate semantic hard negatives generated via VLM rollouts. The resulting candidate set spans the full action space, requiring the critic to distinguish Optimal actions from Suboptimal alternatives, Semantic Distractors, and Unrelated Errors.

\textbf{Hierarchical Semantic Taxonomy.} As illustrated in Figure~\ref{fig:motivation}, GUI affordance manifests as a continuous spectrum rather than a binary state. To capture this topology, we define a four-level semantic taxonomy as evaluation anchors on the continuous affordance space: (1) \textit{Optimal}: The most efficient action strictly aligned with the user intent. (2) \textit{Suboptimal}: Actions that functionally advance the task but incur unnecessary costs or redundancy. (3) \textit{Semantic Distractor}: Actions that are logically related to the intent domain but functionally incorrect. (4) \textit{Unrelated Error}: Actions with no logical or functional connection to the objective.

Annotation follows a two-level decision tree anchored on execution-verifiable judgments: first judge whether the action advances the task (separating Optimal/Suboptimal from Distractor/Unrelated), then subdivide within each group (see Appendix~\ref{app:bbench} for full guidelines). The key boundary---Suboptimal vs.\ Distractor---is anchored on \textit{functional progress}, a criterion verifiable through execution. Importantly, these four levels serve solely as evaluation anchors, not training labels. BBCritic is trained with only binary supervision (Sec.~\ref{twostage}); BBBench tests whether the learned continuous scores recover this ordering without explicit multi-level signal.

\textbf{Statistics.}
As summarized in Table~\ref{tab:dataset-comparison}, BBBench contains 18,192 human-verified samples (full label distribution in Appendix Table~\ref{tab:label_dist}). The natural rarity of Suboptimal samples reflects an asymmetry: most GUI pages offer few redundant-but-valid paths, making the Suboptimal--Distractor boundary the hardest to resolve. To ensure evaluation integrity, BBBench is constructed from MobiBench, which has no sample overlap with the training sources (AndroidControl and GUI Odyssey). Unlike previous datasets limited to binary labels or sparse candidates ($<2$ actions/sample), BBBench provides the first testbed for evaluating page-wise ranking across the full affordance spectrum.

\subsection{Evaluation Metrics}
We employ two complementary metrics to assess whether the critic captures the continuous ranking structure of GUI operations across all four functional levels.

\textbf{Ranking Quality.} Standard accuracy metrics overlook the granular distinctions between non-optimal candidates. To quantify the model's capability in hierarchical semantic modeling—a prerequisite for efficient Test-Time Scaling—we employ Normalized Discounted Cumulative Gain (NDCG). The exponential gain function is defined as $rel(a) = 2^{\ell(a)} - 1$, where $\ell(a) \in \{3, 2, 1, 0\}$ corresponds to the taxonomy tiers. This formulation imposes a strict penalty on ranking suboptimal actions above optimal ones. Finally, NDCG@K and NDCG@All are reported to evaluate both local retrieval utility and global ranking capability.

\textbf{Topology Consistency.} To verify the monotonicity of the Affordance Space ($s_{\text{opt}} > s_{\text{sub}} > s_{\text{dis}} > s_{\text{unr}}$), we introduce Pairwise Preference Accuracy (PPA). PPA measures the decision boundary resolution across adjacent semantic tiers. It is formally defined as the probability that a semantically superior action $u$ is scored higher than an inferior action $v$: $\text{PPA}(\mathcal{U}, \mathcal{V}) = \mathbb{E}_{u \in \mathcal{U}, v \in \mathcal{V}} \left[\mathbb{I}(s_u > s_v) + 0.5\mathbb{I}(s_u = s_v)\right]$ where $(\mathcal{U}, \mathcal{V})$ represents sets of actions from adjacent tiers (e.g., Optimal vs. Suboptimal). Intuitively, PPA measures whether the model can correctly order a random pair of actions from adjacent semantic levels---a score of 50\% indicates chance-level discrimination, while 100\% means perfect boundary resolution. This metric directly tests whether the learned scoring function preserves the expected monotonic ordering across semantic tiers. The full evaluation protocol, including scoring functions for binary baselines and general VLMs, is detailed in Appendix~\ref{app:eval}.

\begin{table}[t]
  \vspace{-5pt}
  \centering
  \caption{Comparison of Critic Models in TTS settings on AndroidControl and GUI Odyssey. Values after (+) denote absolute performance gains, which we prioritize to mitigate the impact of policy rollout variance. Results marked with * are our reproductions; other baselines and their corresponding gains are reported from original literature.}
  \label{tab:mobile_navi}
  
  \small
  \setlength{\tabcolsep}{4pt}

  \resizebox{\textwidth}{!}{%
  \begin{tabular}{lcllllll}
    \toprule
    \multirow{2}{*}{\textbf{Method}} & 
    \multirow{2}{*}{\textbf{Rollout}} & 
    \multicolumn{2}{c}{\textbf{{AndroidControl High}}} & 
    \multicolumn{2}{c}{\textbf{{AndroidControl Low}}} & 
    \multicolumn{2}{c}{\textbf{{GUI Odyssey}}} \\ 
    
    \cmidrule(lr){3-4} \cmidrule(lr){5-6} \cmidrule(lr){7-8}
    
     & & SR & TR & SR & TR & SR & TR \\ 
    \midrule
    
    \multicolumn{8}{c}{\textit{Generative GUI Agent}} \\ 
    \midrule
    OS-Atlas~\cite{wu2024atlas} & - & 29.8 & 57.4 & 50.9 & 73.0 & 27.0 & 60.4 \\
    SeeClick~\cite{cheng2024seeclick}      & - & 59.1 & 82.9 & 75.0 & 93.0 & 53.9 & 71.0 \\
    UI-R1~\cite{lu2025ui}         & - & 45.4 & 57.9 & 66.4 & 79.2 & 32.5 & 52.2 \\
    GUI-R1~\cite{luo2025gui}        & - & 46.6 & 58.0 & 64.4 & 83.7 & 41.3 & 54.8 \\ 
    
    \midrule
    \multicolumn{8}{c}{\textit{Test-time Scaling GUI Agent}} \\ 
    \midrule
    
    Qwen2.5VL-7B*          & -  & 57.9 & 74.2 & 81.2 & 93.0 & 47.2 & 75.7 \\
    \hspace{1em} + UI-Genie \cite{xiao2025ui} & 10 & 65.2 \gain{0.3} & - & - & - & - & - \\
    \hspace{1em} + GAIA \cite{wang2026gaia}    & 8  & 63.8 \gain{3.2} & 84.4 \gain{1.4} & 81.8 \steady  & 94.6 \gain{0.2} & 44.8 \gain{4.6}  & 58.1 \gain{5.4} \\
    
    \rowcolor{rowgray}
    \hspace{1em} \textbf{+ BBCritic-3b}* & 8  & 66.5 \gain{8.6} & 78.7 \gain{4.5} & 86.9 \gain{5.7} & 95.2 \gain{2.2} & 62.3 \gain{15.1} & 83.6 \gain{7.9} \\
    \rowcolor{rowgray}
    \hspace{1em} \textbf{+ BBCritic-7b}* & 8  & 67.7 \gain{9.8} & 79.4 \gain{5.2} & 87.4 \gain{6.2} & 95.4 \gain{2.4} & 62.5 \gain{15.3} & 83.7 \gain{8.0} \\ 
    
    \addlinespace[6pt] 
    
    UITARS-1.5-7B*         & -  & 56.6 & 71.9 & 69.2 & 81.4 & 50.5 & 74.9 \\
    \hspace{1em} + GUI-Shepherd \cite{chen2025gui} & 3  & 65.8 \gain{3.5}  & 81.7 \gain{4.3} & 87.0 \steady    & 95.4 \gain{0.1} & -            & - \\
    \hspace{1em} + GAIA \cite{wang2026gaia}        & 8  & 65.6 \gain{7.4}  & 84.6 \gain{4.5} & 79.2 \gain{7.0}  & 90.1 \gain{3.7} & 50.2 \gain{17.3} & 80.2 \gain{9.1} \\
    
    \rowcolor{rowgray}
    \hspace{1em} \textbf{+ BBCritic-3b}* & 8 & 69.2 \gain{12.6} & 81.3 \gain{9.4} & 81.4 \gain{12.2} & 90.6 \gain{9.2} & 64.5 \gain{14.0} & 84.9 \gain{10.0} \\
    \rowcolor{rowgray}
    \hspace{1em} \textbf{+ BBCritic-7b}* & 8 & 70.6 \gain{14.0} & 81.6 \gain{9.7} & 81.7 \gain{12.5} & 90.9 \gain{9.5} & 66.4 \gain{15.9} & 85.9 \gain{11.0} \\
    
    \bottomrule
  \end{tabular}}
  \vspace{-5pt}
\end{table}

\begin{table}[t]
  \centering
  \caption{Performance comparison on ScreenSpot V2 across Mobile, Desktop, and Web domains. Following the notation in Table~\ref{tab:mobile_navi}, we report absolute performance gains (marked with +) over the baseline and denote reproduced models with *.}
  \label{tab:dataset-screenspotv2}
  
  \small
  \setlength{\tabcolsep}{3pt}

  \resizebox{\textwidth}{!}{%
  \begin{tabular}{lclllllll}
    \toprule
    \multirow{2}{*}{\textbf{Method}} & 
    \multirow{2}{*}{\textbf{Rollout}} & 
    \multicolumn{2}{c}{\textbf{Mobile}} & 
    \multicolumn{2}{c}{\textbf{Desktop}} & 
    \multicolumn{2}{c}{\textbf{Web}} & 
    \multirow{2}{*}{\textbf{Avg.}} \\
    
    \cmidrule(lr){3-4} \cmidrule(lr){5-6} \cmidrule(lr){7-8}
    
     & & Text & Icon & Text & Icon & Text & Icon & \\
    \midrule
    
    GPT-4o \cite{openai_gpt5_2026}       & - & 30.5 & 23.2 & 20.6 & 19.4 & 11.1 & 7.8  & 18.8 \\
    OS-Atlas~\cite{wu2024atlas} & - & 93.0 & 72.9 & 91.8 & 62.9 & 90.9 & 74.3 & 82.5 \\
    SeeClick \cite{cheng2024seeclick}     & - & 78.0 & 52.0 & 72.2 & 30.0 & 55.7 & 32.5 & 53.4 \\
    Aguvis \cite{xu2024aguvis}        & - & 95.6 & 77.7 & 93.8 & 67.1 & 88.3 & 75.2 & 84.4 \\
    
    \midrule
    \multicolumn{9}{c}{\textit{Test-time Scaling GUI Agent}} \\
    \midrule
    
    Qwen2.5VL-7B*          & - & 89.5 & 68.2 & 80.9 & 41.4 & 70.0 & 50.7 & 66.8 \\
    
    \hspace{1em} + GAIA \cite{wang2026gaia}  & 8 & 89.7 \gain{4.9} & 68.2 \gain{8.5} & 78.9 \gain{6.8} & 54.3 \gain{2.2} & 76.9 \gain{7.7} & 51.2 \gain{7.9} & 71.1 \gain{6.1} \\
    
    \rowcolor{rowgray}
    \hspace{1em} \textbf{+ BBCritic-3b}* & 8 
    & 93.5 \gain{4.0} 
    & 74.1 \gain{5.9} 
    & 85.0 \gain{4.0} 
    & 51.1 \gain{9.7} 
    & 84.6 \gain{14.6} 
    & 70.6 \gain{19.9} 
    & 76.8 \gain{10.0} \\ 
    
    \rowcolor{rowgray}
    \hspace{1em} \textbf{+ BBCritic-7b}* & 8 
    & 94.5 \gain{5.0} 
    & 76.4 \gain{8.2} 
    & 88.1 \gain{7.2} 
    & 60.7 \gain{19.3} 
    & 89.4 \gain{19.4} 
    & 76.7 \gain{25.9} 
    & 81.0 \gain{14.2} \\ 
    
    \bottomrule
  \end{tabular}}
  \vspace{-10pt}
\end{table}

\section{Experiments}

\subsection{Implementation Details}

\textbf{Model and Hyperparameters.} We instantiate our critic models using Qwen2.5-VL-Instruct (3B and 7B)~\cite{bai2025qwen2} as backbones. Fine-tuning is conducted via LoRA with $r=16$. We optimize using AdamW with a learning rate of $5 \times 10^{-5}$, and set the contrastive temperature $\tau$ to $0.02$. Stage~1 is trained for 1 epoch and Stage~2 for 2 epochs (see Table~\ref{tab:hyperparams} in Appendix~\ref{app:training} for the full configuration).

\textbf{Curation Settings.} Following the pipeline in Sec.~\ref{twostage}, we utilize OmniParserV2~\cite{yu2025omniparser} for layout parsing. For the hard negative mining rollout, we employ a Qwen2.5-VL-7B policy model with a sampling size of $N=64$.

\textbf{Data Source and Scale.} The training data is derived from the training splits of AndroidControl~\cite{li2024effects} and GUIOdyssey~\cite{lu2025guiodyssey}. Stage~1 contains 191K samples (AndroidControl 89K + GUIOdyssey 102K) with an average of ${\sim}58$ layout-parsed negatives per sample. Stage~2 contains 190K samples (89K + 101K) with ${\sim}25$ VLM-rollout hard negatives per sample. All experiments are conducted under a weakly supervised setting without any additional annotation or extra data filtering.
\subsection{Main Results}

\subsubsection{Performance on Semantic Modeling}
As shown in Table~\ref{tab:critic-performance}, our method achieves the strongest aggregate performance among open-source models on \textsc{BBBench} across both NDCG and PPA metrics. Notably, binary critics fine-tuned from Qwen2.5-VL-7B (e.g., GAIA~\cite{wang2026gaia}, GUI-Critic-R1~\cite{wanyan2025look}) consistently fall below their backbone on \textsc{BBBench}, despite extensive critic-specific training. This counter-intuitive result empirically supports our analysis of Affordance Collapse: binary objectives compress the action topology into scalar probabilities, conflating suboptimal yet valid attempts with semantic distractors. In contrast, our metric learning formulation recovers this semantic hierarchy, enabling precise quality ranking. We note that BBCritic-3B slightly outperforms 7B on $\text{PPA}_{\text{opt-sub}}$ (80.99 vs.\ 78.72), while 7B leads on all other metrics. We attribute this minor gap to larger backbones being more tolerant of redundant-but-valid paths, occasionally conflating Optimal and Suboptimal at the finest boundary. On one GPU, BBCritic-3B/7B take 1.56s/2.17s per step (CoT-prompted Qwen2.5-VL-7B: 5.89s; GPT-5: 10.8s per item; Gemini 3 Pro: 32.1s per item). A page-level case study illustrating these score distributions on a representative episode is provided in Appendix~\ref{app:qualitative}.
\vspace{-5pt}
\subsubsection{Offline TTS and Generalization Testing}

\textbf{In-Domain and Cross-Task Consistency.} As shown in Table~\ref{tab:mobile_navi}, our 3B model surpasses 7B SOTA baselines on the in-domain high-level navigation dataset. More importantly, this performance gain transfers consistently to the AndroidControl-Low dataset. This validates our Functional Equivalence Hypothesis: although high-level and low-level instructions differ linguistically, the optimal action sequences share an equivalent functional logic. By modeling the underlying semantic alignment rather than surface-level pattern matching, our critic generalizes consistently across instruction granularity.

\textbf{Cross-Platform Invariant Semantics.} Tables~\ref{tab:dataset-screenspotv2}~and~\ref{tab:generalization-comparison} report the results on cross-platform benchmarks (ScreenSpotV2~\cite{wu2024atlas} and Mind2Web~\cite{deng2023mind2web}). Despite never being exposed to Web or Desktop data during training, our model demonstrates strong zero-shot transferability, achieving a 14.2\% improvement over the baseline on ScreenSpotV2. We attribute this generalization to the learning of invariant action semantics. The underlying interaction logic (e.g., ``search,'' ``confirmation'') is shared across platforms. This suggests that our contrastive objective captures the core semantics of GUI interactions, allowing the critic to generalize beyond platform-specific visual differences. On the Mobile-Icon split of ScreenSpotV2, our gain (+8.2) is marginally below GAIA's (+8.5), likely because our semantic embedding compresses fine-grained positional information; this gap closes on Desktop/Web where semantic reasoning dominates over precise grounding.

\subsubsection{Online Evaluation in AndroidWorld}
Table~\ref{tab:androidworld} presents the results in the dynamic AndroidWorld~\cite{rawles2024androidworld} environment. Following the formulation in Sec.~\ref{motivation}, TTS requires the critic to rank $N$ candidates and select $a_t^* = \arg\max s_t^i$. Binary critics provide only accept/reject signals and must approximate this through multi-turn rejection sampling—a procedure that fails when all candidates are rejected (false negatives), whereas our continuous scores fit ranking-based selection directly. Our method not only improves the overall success rate but also significantly outperforms binary models when they are adapted to the ranking setting. This indicates that binary classifiers are brittle when facing ambiguous candidates in real-world rollouts, whereas our continuous scores provide a more stable ranking for action selection.

\begin{table}[!t]
  \centering
  \caption{Cross-scenario generalization on Mind2Web. The ``Website Data'' column indicates whether web-domain data was used during training. Following the notation in Table~\ref{tab:mobile_navi}, gains marked with + are absolute improvements over the baseline; * denotes our reproductions.}
  \label{tab:generalization-comparison}
  \small 
  \setlength{\tabcolsep}{4pt} 

  \begin{tabular}{lcc}
    \toprule
    \textbf{Method} & \textbf{Website Data} & \textbf{MM-Mind2Web} \\
    \midrule
    
    SeeAct \cite{zheng2024gpt}      & \cmark & 63.0 \\
    Aguvis \cite{xu2024aguvis}      & \cmark & 57.2 \\
    \midrule
    
    \multicolumn{3}{c}{\textit{Test-Time Scaling GUI Agent}} \\
    \midrule
    
    Qwen2.5VL-7B* & - & 57.5 \\
    
    \rowcolor{rowgray}
    + BBCritic-7b* & \xmark & 61.9 \gain{4.4} \\
    
    \rowcolor{rowgray}
    + BBCritic-7b* & \cmark & 64.8 \gain{7.3} \\
    \bottomrule
  \end{tabular}
  \vspace{-4pt}
\end{table}

\begin{table}[!t]
  \centering
  \caption{Online evaluation on AndroidWorld under Test-Time Scaling settings. We report Success Rate and absolute gains (+) over the baseline. All baselines are reproduced for fair comparison.}
  \label{tab:androidworld}
  
  \small
  \setlength{\tabcolsep}{6pt}

  \begin{tabular}{llc}
    \toprule
    \textbf{Method} & \textbf{Critic Method} & \textbf{Success Rate} \\
    \midrule
    
    Qwen2.5-VL & - & 25.5 \\
    \midrule
    
    \hspace{1em} + GUI-Critic-R1 & Multi-turn & 29.4 \gain{3.9} \\
    \hspace{1em} + GAIA          & Multi-turn & 27.2 \gain{1.7} \\
    \hspace{1em} + GAIA          & Ranking    & 26.8 \gain{1.3} \\
    
    \rowcolor{rowgray}
    \hspace{1em} \textbf{+ BBCritic-3B} & Ranking & 29.8 \gain{4.3} \\
    \rowcolor{rowgray}
    \hspace{1em} \textbf{+ BBCritic-7B} & Ranking & 30.2 \gain{4.7} \\
    
    \bottomrule
  \end{tabular}
  \vspace{-10pt}
\end{table}

\begin{table}[!t]
  \centering
  \caption{Ablation study on the two-stage curriculum (BBCritic-3B). We report Success Rate (SR) and decision Margin on AndroidControl-High and GUI Odyssey.}
  \label{tab:ablation-stage}
  \small
  \setlength{\tabcolsep}{6pt}

  \begin{tabular}{lcccc}
    \toprule
    \multirow{2}{*}{\textbf{Model}} & 
    \multicolumn{2}{c}{\textbf{\shortstack{AndroidControl\\High}}} & 
    \multicolumn{2}{c}{\textbf{\shortstack{GUI\\Odyssey}}} \\
    
    \cmidrule(lr){2-3} \cmidrule(lr){4-5}
    
     & \textbf{SR} & \textbf{Margin} & \textbf{SR} & \textbf{Margin} \\
    \midrule
    
    \rowcolor{rowgray}
    \textbf{Ours} & \textbf{66.5} & \textbf{0.99} & \textbf{62.3} & \textbf{1.35} \\
    
    w/o Stage 1   & 64.3 & 0.53 & 61.3 & 0.87 \\
    w/o Stage 2   & 61.3 & 0.72 & 58.8 & 1.12 \\
    
    \bottomrule
  \end{tabular}
  \vspace{-10pt}
\end{table}

\subsection{Ablation Study \& Analysis}

We ablate four key design choices. Full details and additional analyses are provided in the Appendix.

\textbf{Two-Stage Curriculum.} As shown in Table~\ref{tab:ablation-stage}, removing Stage~1 halves the decision margin (0.99$\to$0.53 on AndroidControl), while removing Stage~2 drops the success rate by 5.2\%. This confirms that Stage~1 establishes the \textit{global semantic topology} and Stage~2 sharpens \textit{local decision boundaries}---both stages are complementary and necessary.

\begin{figure}[!t]
  \begin{center}
    \centerline{\includegraphics[width=0.6\textwidth]{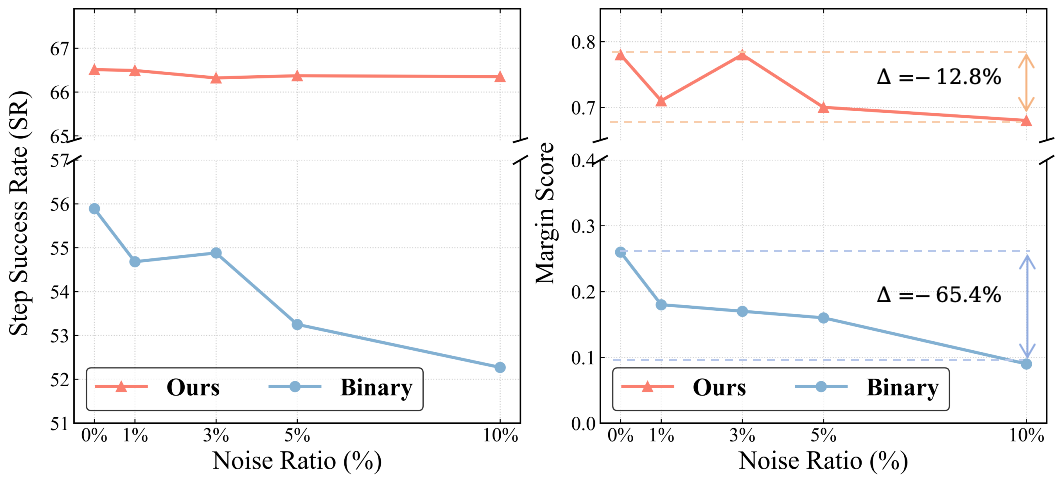}}
    \caption{
    Impact of label noise on AndroidControl TTS performance. Binary critics collapse on both margin and SR, while BBCritic remains stable on both.
    }
    \label{fig:noise_analysis}
  \end{center}
  \vspace{-15pt}
\end{figure}

\textbf{Robustness to Semantic Label Noise.} We evaluate robustness by introducing random label flipping during training (Figure~\ref{fig:noise_analysis}). Binary models suffer rapid decision margin collapse, indicating overfitting to noisy artifacts. Our method maintains distinct score margins with minimal degradation, confirming that the contrastive objective prioritizes \textit{global ranking order} over fitting individual noisy labels.

\textbf{Hard Negative Source \& Data Scale.} VLM-rollout negatives improve Recall@1 from 60.5 to 66.2 over model-ranked mining (Table~\ref{tab:hard_neg}); ranking capability remains stable at 25\% data, with additional samples mainly sharpening the Sub--Dis boundary (48.2$\to$51.2 PPA, Table~\ref{tab:data_scale}). Full analysis is deferred to Appendix~\ref{app:ablation}.

\textbf{InfoNCE vs.\ Binary Cross-Entropy.} Under identical data and varying negative density, BCE collapses (margins fall below zero at high density) while InfoNCE improves steadily, confirming that contrastive learning converts dense negatives into useful signals rather than noise (Figure~\ref{fig:neg_num_analysis}, full discussion in Appendix~\ref{app:ablation}).

\section{Conclusion}

This paper recasts GUI critique from binary classification into continuous metric learning over a shared Affordance Space. Anchored on the Functional Equivalence Hypothesis, BBCritic operationalizes this view through a two-stage contrastive curriculum that resolves the affordance collapse of binary objectives. To probe this finer structure, we release BBBench—a dense-candidate benchmark with a four-level taxonomy—on which BBCritic-3B surpasses 7B-parameter binary critics without additional annotations and transfers zero-shot across platforms. These results support a methodological shift: GUI critique is fundamentally a metric-learning problem, not a classification one.

\textbf{Limitations.} BBCritic is currently validated only under Test-Time Scaling. Critic models naturally extend to broader settings—training data curation, online RL reward modeling, agent self-correction, human-in-the-loop studies—which we leave to future work due to infrastructure constraints rather than methodological barriers. Our training data also spans only mobile platforms (AndroidControl and GUI Odyssey); while zero-shot transfer to Desktop/Web is encouraging, multi-platform training would likely further sharpen platform-specific grounding. Finally, BBBench's coverage of rare categories such as Suboptimal samples is bounded by the human verification cost of execution-grounded annotation; scaling such fine-grained labels at low cost remains an open community challenge. Additional considerations are discussed in Appendix~\ref{app:impact}.

\bibliography{ref}

\newpage
\appendix

\section{BBBench Construction Details} \label{app:bbench}

\subsection{Annotation Guidelines}

We adopt a structured two-level decision tree to ensure consistent annotation across annotators. The full procedure is as follows:

\textbf{Level 1 --- Functional Progress.} For each candidate action, the annotator first answers: \textit{``Does executing this action positively advance the user's task toward completion?''} This binary judgment separates actions into a \textbf{Positive group} (task-advancing) and a \textbf{Negative group} (non-advancing). The key criterion is whether the action brings the agent measurably closer to the goal state, which can be verified through execution in a virtual machine.

\textbf{Level 2 --- Subdivision.}
\begin{itemize}
  \item \textbf{Within the Positive group:} The annotator judges whether the action follows the most efficient path. If yes $\to$ \textit{Optimal} (relevance level $\ell=3$); if the action advances the task but introduces unnecessary steps or redundancy $\to$ \textit{Suboptimal} ($\ell=2$).
  \item \textbf{Within the Negative group:} The annotator judges whether the action is semantically related to the instruction domain. If the target element belongs to the same functional category as the instruction (e.g., clicking ``Cart'' when the instruction is ``Buy'') $\to$ \textit{Semantic Distractor} ($\ell=1$); if the action has no logical connection $\to$ \textit{Unrelated Error} ($\ell=0$).
\end{itemize}

The critical boundary between Suboptimal and Semantic Distractor is anchored on \textit{functional progress}: a Suboptimal action must demonstrably advance the task (even if inefficiently), while a Semantic Distractor may appear related but fails to make progress. The relevance levels $\ell \in \{3, 2, 1, 0\}$ defined here directly feed into the NDCG and PPA metrics introduced in Sec.~\ref{sec:bbbench}.

\subsection{Annotation Pipeline}

Our annotation process follows a multi-stage quality assurance protocol:

\begin{enumerate}
  \item \textbf{Independent Annotation.} Three trained annotators independently label each candidate action using the decision tree above.
  \item \textbf{Majority Voting.} For each action, the final label is determined by majority vote (at least 2 out of 3 agree).
  \item \textbf{Virtual Machine Execution Verification.} Ambiguous cases---particularly those on the Suboptimal--Distractor boundary---are resolved by executing the action in a mobile device emulator and observing whether the task state progresses.
  \item \textbf{Human Review.} A senior annotator reviews all cases where the three initial annotators disagreed, making the final determination based on the execution evidence.
\end{enumerate}

\textbf{Annotator Background.} All annotators have prior experience in GUI navigation data annotation (at least 3 months). Before the annotation campaign, annotators completed a training session with 50 calibration examples covering all four taxonomy levels, discussing edge cases and aligning on the decision tree criteria.

\textbf{Why Execution-Grounded Verification.} The most subjective step in our taxonomy is the Suboptimal--Distractor boundary, where annotators must judge whether an action makes \textit{functional progress} toward the goal. Rather than relying on annotator intuition alone, our pipeline anchors this boundary on objective execution evidence: any candidate that three annotators do not unanimously agree on is replayed in a mobile emulator, and the post-action UI state determines the final label. This design moves the Suboptimal--Distractor decision from a matter of subjective interpretation to one of observable state change, which is the primary mechanism through which BBBench achieves consistent labels at the most ambiguous boundary.

\subsection{Label Distribution and Statistics}

\begin{table}[h]
  \centering
  \caption{Label distribution in BBBench. The low proportion of Suboptimal samples (3.6\%) reflects the natural rarity of redundant-but-valid interaction paths on typical GUI pages.}
  \label{tab:label_dist}
  \small
  \begin{tabular}{lrr}
    \toprule
    \textbf{Level} & \textbf{Count} & \textbf{Proportion} \\
    \midrule
    Optimal & 1,612 & 8.9\% \\
    Suboptimal & 652 & 3.6\% \\
    Semantic Distractor & 2,871 & 15.8\% \\
    Unrelated Error & 13,057 & 71.8\% \\
    \midrule
    \textbf{Total} & \textbf{18,192} & \textbf{100\%} \\
    \bottomrule
  \end{tabular}
\end{table}

BBBench averages 30.78 candidate actions per page, providing a dense evaluation space that is an order of magnitude larger than existing benchmarks ($<2$ actions/sample). The high proportion of Unrelated Errors (71.8\%) reflects the reality of GUI pages: most interactive elements are irrelevant to any given instruction. The scarcity of Suboptimal samples (3.6\%) represents a natural asymmetry---most pages offer few redundant-but-valid paths---and makes the Suboptimal--Distractor boundary the hardest to evaluate.

\textbf{Coverage and Diversity.} BBBench is constructed strictly as an evaluation benchmark and does not participate in any training. The 18{,}192 candidates are organized into 591 task-step pages spanning 102 distinct tasks across more than 50 Android applications, covering a broad range of categories such as productivity (Calendar, Files, Joplin), media (Audio Recorder, ESPN, BBC), social (Discord, Instagram, Gmail), travel (Maps, Booking.com, DoorDash), and learning (Coursera, Duolingo). User instructions average 84 words (median 52), reflecting realistic high-level task descriptions rather than short atomic commands. The candidate action set is dominated by \texttt{click} actions (72.6\%), with the remaining 27.4\% covering scroll variants (13.2\%), \texttt{type} (3.8\%), \texttt{open\_app} (3.3\%), \texttt{navigate\_back} (3.3\%), and \texttt{terminate} (3.3\%). This composition matches the action distribution observed in mobile navigation and ensures that BBBench evaluates critics across both interaction-heavy and command-style operations.

\subsection{Data Leakage Check}

BBBench is constructed from the MobiBench~\cite{im2025modular} dataset, while all training data for BBCritic is derived exclusively from the training splits of AndroidControl~\cite{li2024effects} and GUIOdyssey~\cite{lu2025guiodyssey}. MobiBench was constructed as an offline evaluation benchmark with documented disjointness from these training sources, and we adopt their disjointness guarantee: no episode, screenshot, or instruction in BBBench appears in our training data. This ensures that BBBench measures genuine generalization rather than memorization.

\section{Background on Universal Multimodal Retrieval} \label{app:umr}

Our framework draws on Universal Multimodal Retrieval (UMR), where the goal is to embed queries and documents---which may consist of any combination of text and images---into a shared dense space for cross-modal matching. This appendix surveys the UMR foundations and the more recent VLM-as-Encoder paradigm, distills the findings that motivate our design, and clarifies how BBCritic both builds on and departs from these works.

\textbf{Universal Multimodal Retrieval.} The unifying objective of UMR is to learn a shared embedding space in which heterogeneous queries (text, image, or mixtures) can be matched against arbitrary documents through cosine similarity. This paradigm originates with CLIP~\cite{radford2021learning}, which trains separate text and image encoders on web-scale image-caption pairs via a contrastive loss, yielding strong zero-shot retrieval and classification. ALIGN~\cite{jia2021scaling} pushed this to billion-scale noisy alt-text data, showing that scale compensates for label quality. SigLIP~\cite{zhai2023sigmoid} replaced the softmax contrastive loss with a sigmoid pairwise loss for more stable scaling. BLIP-2~\cite{li2023blip2} bridged frozen vision encoders and frozen large language models with a lightweight Q-Former, anticipating the now-dominant practice of repurposing pre-trained VLMs as multimodal representation modules. Collectively, these works establish dense embeddings as a versatile primitive for cross-modal retrieval.

\textbf{VLM-as-Encoder Paradigm.} A more recent line of work goes one step further by repurposing pre-trained Vision-Language Models (VLMs) directly as multimodal encoders, leveraging the rich cross-modal alignment learned during generative pre-training. Methodologically, these approaches share a common recipe: feed the multimodal input through the VLM, extract the hidden state of the final (typically EOS) token as a dense embedding, and fine-tune the model with a contrastive objective such as InfoNCE on (query, document) pairs. GME~\cite{zhang2024gme} instantiates this recipe on Qwen2-VL and achieves strong performance across text-only, image-only, and mixed-modality retrieval. VLM2Vec~\cite{jiang2024vlm2vec} adopts the same architectural choice and adds task-specific instruction tuning. E5-V~\cite{jiang2024e5v} bridges the modality gap with a prompt-based scheme that allows training on text pairs alone while still embedding images. LLaVE~\cite{lan2025llave} improves embedding discriminability through hardness-weighted contrastive learning. NV-Embed~\cite{lee2024nvembed} systematically compares pooling strategies and reports that last-token and latent-attention pooling generally outperform mean pooling. ColPali~\cite{faysse2024colpali} extends the paradigm to page-level visual document retrieval. Together, these works show that pre-trained VLMs, originally optimized for autoregressive generation, can be effectively adapted into bidirectional dense encoders without architectural overhaul.

\textbf{Findings That Motivate Our Design.} Three observations from this literature directly shape BBCritic. First, the \textit{last-token pooling} strategy~\cite{zhang2024gme, lee2024nvembed} consistently produces strong embeddings from generative VLMs without architectural changes, making it the natural default for adapting Qwen2.5-VL into a dense encoder. Second, \textit{hard negatives substantially shape the embedding geometry}~\cite{lan2025llave}: when many semantically similar candidates compete with a positive, the contrastive loss must explicitly leverage their relative confusion to avoid representation collapse---this aligns with our InfoNCE-over-page-candidates objective (Sec.~\ref{hypos}). Third, \textit{page-level retrieval is feasible}~\cite{faysse2024colpali}: ColPali shows that VLM-based encoders can index visual documents at page granularity, supporting our choice of treating each GUI screen as a self-contained candidate set rather than flattening across the full corpus.

\textbf{Connection to BBCritic.} BBCritic inherits the core implementation recipe of VLM-as-Encoder UMR---last-token hidden state as the embedding, InfoNCE as the contrastive loss, Siamese encoder for both modalities---but addresses a distinct objective. UMR seeks to retrieve documents that semantically match a query across modalities; BBCritic seeks to rank candidate GUI actions by their functional alignment with a user instruction. The shared insight is that pre-trained VLMs serve as powerful dense cross-modal encoders. The key conceptual departure is that we treat user instructions and GUI actions as two expressions of the same underlying \textit{affordance} (the Functional Equivalence Hypothesis, Sec.~\ref{hypos}), rather than as a heterogeneous query-document pair. Operationally, we further optimize for ranking quality across an entire candidate set on a single page rather than for binary relevance against a global corpus---this page-level structure mirrors ColPali's granularity choice, but applied to dynamic GUI affordances rather than static document content.

\section{Training Details} \label{app:training}

\subsection{Hyperparameters}

Table~\ref{tab:hyperparams} lists the complete training configuration used in all experiments.

\begin{table}[h]
  \centering
  \caption{Complete training hyperparameters for BBCritic.}
  \label{tab:hyperparams}
  \small
  \begin{tabular}{ll}
    \toprule
    \textbf{Parameter} & \textbf{Value} \\
    \midrule
    Backbone & Qwen2.5-VL-Instruct (3B / 7B) \\
    Fine-tuning method & LoRA ($r=16$, $\alpha=64$) \\
    Optimizer & AdamW \\
    Learning rate & $5 \times 10^{-5}$ \\
    LR schedule & Linear with warmup ratio $0.03$ \\
    Precision & bfloat16 \\
    Contrastive temperature $\tau$ & 0.02 \\
    Stage 1 epochs & 1 \\
    Stage 2 epochs & 2 \\
    Batch size & 1 per GPU $\times$ 8 GPUs (effective 8) \\
    Hard negatives per sample & up to 64 (both stages) \\
    \bottomrule
  \end{tabular}
\end{table}

\subsection{Training Data Construction}

\textbf{Stage 1: Layout-Parsed Negatives.} We construct 191,230 training samples from AndroidControl (89K) and GUIOdyssey (102K). For each sample, we parse the full UI layout with OmniParserV2~\cite{yu2025omniparser} to extract all interactable elements as negatives, yielding an average of ${\sim}58$ negative candidates per sample. Non-click actions (e.g., scroll, type) are supplemented via rule-based generation to ensure functional completeness.

\textbf{Stage 2: VLM-Rollout Hard Negatives.} We construct 190,072 samples (AndroidControl 89K + GUIOdyssey 101K). For each step, we employ a Qwen2.5-VL-7B policy model to generate $N=64$ sampled rollouts (temperature 1.0). Each sampled action is then verified against the ground-truth next action at the same step; actions that diverge from the ground truth---i.e., would not advance the task at this step---are harvested as semantic hard negatives. This yields an average of ${\sim}25$ hard negatives per sample. Combined with layout-parsed negatives retained from Stage 1, each sample contains a comprehensive candidate set spanning the full difficulty spectrum.

\subsection{Set-of-Mark Prompting} \label{app:som}

A critical challenge in GUI evaluation is the high visual density of interface elements. Relying on raw pixels forces the model to simultaneously perform \textit{visual grounding} (locating the element) and \textit{semantic reasoning} (evaluating its utility), often diluting the model's attention. To mitigate this, we integrate Set-of-Mark style visual prompting~\cite{yang2023set} during the action encoding phase. By visually superimposing a red circle marker onto the target click position of candidate action $a^i$, we explicitly direct the model's attention to the relevant region. This design decouples the low-level burden of localization from high-level reasoning, allowing the model to focus on evaluating the semantic match between the marked operation and the instruction.

\textbf{Implementation.} For each candidate action $a^i$ requiring spatial localization (\texttt{click} or \texttt{long\_press}), we render a separate image variant where a red circle marker (radius 25 px, stroke width 8 px) is overlaid at the action's target coordinate. Non-spatial actions (scroll, type, app launch, system buttons) reuse the raw screenshot without any marker. This per-candidate rendering scheme lets the encoder ground each action against the same screen context with the target spatially highlighted, avoiding the indexing ambiguity that arises when many marks coexist on a single shared image.

\section{Extended Ablation Studies} \label{app:ablation}

\subsection{InfoNCE vs.\ Binary Cross-Entropy: Impact of Negative Sample Density}

\begin{figure}[ht]
  \begin{center}
    \centerline{\includegraphics[width=0.6\textwidth]{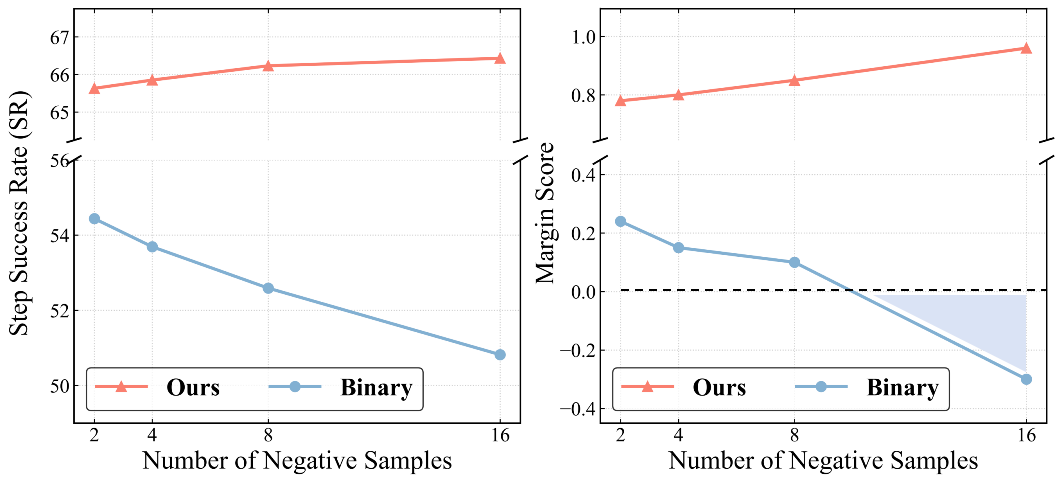}}
    \caption{
    Impact of the number of negative samples on AndroidControl TTS performance. BCE suffers gradient saturation with increasing density, while InfoNCE improves steadily.
    }
    \label{fig:neg_num_analysis}
  \end{center}
\end{figure}

To isolate the contribution of the contrastive objective from data effects, we train BBCritic-3B with the Stage 1 data, varying only the negative candidate density across $\{2, 4, 8, 16\}$ while keeping all other hyperparameters fixed. As shown in Figure~\ref{fig:neg_num_analysis}, the two objectives diverge sharply with increasing density. The binary (BCE) baseline suffers from gradient saturation: its step success rate (SR) drops from 54.4 to 50.8 and its decision margin collapses from $+0.24$ to $-0.30$ at 16 negatives, indicating that dense candidates overwhelm the independent gradient signal. In contrast, InfoNCE improves steadily on both metrics---SR from 65.6 to 66.4 and margin from $+0.78$ to $+0.96$---confirming that contrastive learning converts dense negatives into useful relational signals via implicit hard negative mining.

\subsection{Training Data Scale}

\begin{table}[h]
  \centering
  \caption{Ablation on training data scale (BBCritic-3B, AndroidControl only). Both Stage 1 and Stage 2 data are proportionally sampled. Ranking capability (NDCG@all) is stable even at 25\% data, while full data primarily sharpens the finest Suboptimal--Distractor boundary.}
  \label{tab:data_scale}
  \small
  \setlength{\tabcolsep}{5pt}
  \begin{tabular}{c|ccc|ccc}
    \toprule
    \multirow{2}{*}{\textbf{Data \%}} & \multicolumn{3}{c|}{\textbf{AndroidControl}} & \multicolumn{3}{c}{\textbf{BBBench}} \\
     & Recall@1 & Recall@5 & Margin & NDCG@all & Opt-Sub & Sub-Dis \\
    \midrule
    25\% & 64.2 & 75.0 & 0.764 & 70.3 & 80.6 & 48.2 \\
    50\% & 65.5 & 74.9 & 0.930 & 69.7 & \textbf{81.9} & 49.2 \\
    \rowcolor{rowgray}
    \textbf{100\%} & \textbf{66.5} & \textbf{75.0} & \textbf{1.135} & \textbf{70.5} & 80.9 & \textbf{51.2} \\
    \bottomrule
  \end{tabular}
\end{table}

These results reveal that the ranking capability of BBCritic stems primarily from the contrastive learning objective rather than data volume: NDCG@all is nearly identical at 25\% and 100\% data (70.3 vs.\ 70.5). The benefit of full data is concentrated on the finest semantic boundary---Sub-Dis PPA improves from 48.2 to 51.2---and on the overall decision margin (0.764 $\to$ 1.135, $+48\%$ relative). In other words, more data does not yield more correct rankings, but produces sharper, more confident decisions at fine-grained boundaries.

\subsection{Hard Negative Source}

\begin{table}[h]
  \centering
  \caption{Ablation on Stage 2 hard negative source (BBCritic-3B, AndroidControl only, 16 hard negatives per sample). VLM rollout negatives provide substantially stronger generalization than model-ranked mining.}
  \label{tab:hard_neg}
  \small
  \setlength{\tabcolsep}{5pt}
  \begin{tabular}{l|ccc|ccc}
    \toprule
    \multirow{2}{*}{\textbf{Source}} & \multicolumn{3}{c|}{\textbf{AndroidControl}} & \multicolumn{3}{c}{\textbf{BBBench}} \\
     & Recall@1 & Recall@5 & Margin & NDCG@all & Opt-Sub & Sub-Dis \\
    \midrule
    Model-ranked Mining & 60.5 & 74.7 & 1.014 & 69.8 & 79.3 & 47.5 \\
    \rowcolor{rowgray}
    \textbf{VLM Rollout (Ours)} & \textbf{66.2} & \textbf{75.1} & \textbf{1.124} & \textbf{80.5} & \textbf{79.6} & \textbf{51.2} \\
    \bottomrule
  \end{tabular}
\end{table}

We compare two hard negative sources under identical settings (same Stage 1 checkpoint, same number of hard negatives): (1) \textbf{Model-ranked mining}, which selects top-scored negatives from Stage 1 candidates using the trained model; (2) \textbf{VLM rollout}, which uses actions generated by a separate VLM policy. VLM rollout negatives yield substantially stronger results (Recall@1: 60.5$\to$66.2, +9.4\% relative; NDCG@all: 69.8$\to$80.5, +15.3\% relative). We attribute this to distribution complementarity: model-mined negatives are drawn from the same layout-parsed pool used in Stage 1, providing limited novelty. VLM rollout negatives, by contrast, represent genuine behavioral errors---plausible actions that a capable policy would actually attempt---and thus occupy the semantic ``gray zone'' that requires boundary sharpening.

\subsection{Why InfoNCE over Other Ranking Losses?} \label{app:infonce_vs_pairwise}

Our choice of InfoNCE is constrained by the weak-supervision setting: training data offers only binary correct/incorrect labels per candidate, with no multi-level relevance annotations. Below we examine three families of alternative ranking losses---ordinal regression, listwise losses, and pairwise margin losses---and explain why InfoNCE is the natural fit for this setting.

\textbf{Ordinal regression}~\cite{herbrich1999support} requires explicit multi-level relevance labels per sample (e.g., 4-level annotations). Since our training data contains only binary correct/incorrect labels, ordinal regression is inapplicable without additional annotation---contradicting our weak-supervision constraint.

\textbf{Listwise losses} (e.g., ListNet~\cite{cao2007learning}) define a distribution over all candidates and optimize the KL divergence with the ground-truth distribution. When relevance labels are binary (one positive, rest negative), the ground-truth distribution is a one-hot vector, and the listwise loss reduces to the cross-entropy between the softmax prediction and the one-hot target---which is mathematically equivalent to the InfoNCE formulation. Thus, listwise losses offer no additional benefit over InfoNCE in our setting.

\textbf{Pairwise margin losses}~\cite{burges2005learning} (e.g., hinge loss: $\max(0, m - s^+ + s^-)$) optimize individual positive--negative pairs independently. While applicable under binary supervision, pairwise losses have a structural limitation: each gradient update considers only a single negative, ignoring the relational structure across all candidates in the batch. InfoNCE, by contrast, normalizes over all negatives via softmax, creating an implicit competition that automatically prioritizes hard negatives (those with high $P(a^{k-})$). This global context is particularly valuable in GUI critique, where dozens of candidates of varying difficulty coexist on each page.

In summary, under the binary supervision constraint, ordinal methods are inapplicable, listwise methods are equivalent to InfoNCE, and pairwise methods are a strictly weaker variant. InfoNCE is therefore the natural and optimal choice for our setting.

\section{Qualitative Analysis} \label{app:qualitative}

To complement the aggregate metrics in the main text, we present a single-page case study from \textsc{BBBench} that illustrates (i) what the four-level taxonomy looks like on a real GUI page, and (ii) how the continuous scores produced by BBCritic, the binary critic GAIA, and a strong general VLM (GPT-5) differ on the same set of candidate actions.

\subsection{Per-Page Taxonomy Visualization}

\begin{figure}[!htbp]
  \centering
  \includegraphics[width=0.5\textwidth]{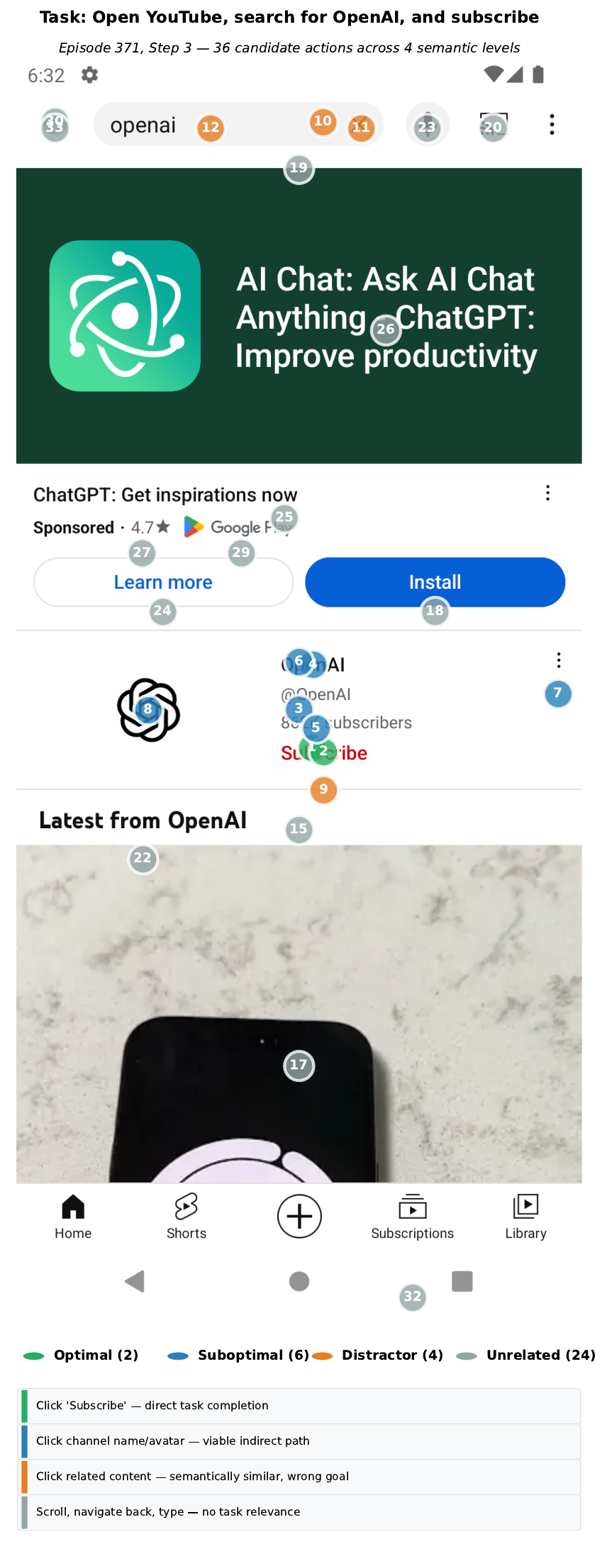}
  \caption{BBBench Episode 371 (Task: \emph{Open YouTube, search for OpenAI, and subscribe}). The 36 candidate actions on this page span all four taxonomy levels: 2 Optimal (green), 6 Suboptimal (blue), 4 Distractor (orange), and 24 Unrelated (gray). Selected examples and their semantic rationale are listed at the bottom.}
  \label{fig:qualitative_taxonomy}
\end{figure}

Figure~\ref{fig:qualitative_taxonomy} shows Episode 371, Step 3 from \textsc{BBBench}, where the user instruction is \emph{``Open YouTube, search for OpenAI, and subscribe''} and the agent is currently on the OpenAI YouTube channel page. Of the 36 candidate actions extracted from this page, two are \textit{Optimal} (the Subscribe button and its edge region), six are \textit{Suboptimal} (related interactions on the same channel that eventually lead to subscription, e.g., opening the channel menu or the OpenAI avatar), four are \textit{Distractor} (visually similar buttons on unrelated UI elements such as the search bar and clear-search icon), and 24 are \textit{Unrelated Errors} (system buttons, app launcher, irrelevant content tiles). This single page already exhibits all four functional levels and reflects the typical density and skew of \textsc{BBBench} (cf.\ Appendix~\ref{app:bbench}): a small number of clearly correct or clearly wrong actions, with a critical ``gray zone'' of Suboptimal and Distractor candidates that requires fine-grained semantic discrimination.

\subsection{Per-Candidate Scoring Comparison}

Table~\ref{tab:qualitative-scoring} compares BBCritic's continuous score against GAIA's binary judgment and GPT-5's prompted scoring on 11 representative candidates from this page (two from each taxonomy level, plus four Unrelated examples to span the long tail).

\begin{table}[!t]
\centering
\caption{Scoring comparison on selected candidates from Episode 371 (Task: \emph{Open YouTube, search for OpenAI, and subscribe}). BBCritic scores decrease monotonically across all four taxonomy levels (8.56 $\to$ 5.28), while GAIA's binary judgment misclassifies all six Suboptimal/Distractor/Unrelated examples as Wrong, and GPT-5 conflates Suboptimal with Distractor and several Unrelated cases under the same ``Semantic Error'' label.}
\label{tab:qualitative-scoring}
\resizebox{\textwidth}{!}{%
\begin{tabular}{clcccccc}
\toprule
\textbf{\#} & \textbf{Action} & \textbf{Label} & \textbf{BBCritic} & \textbf{GAIA Score} & \textbf{GAIA} & \textbf{GPT-5 Score} & \textbf{GPT-5} \\
\midrule
\rowcolor{green!8} 1 & Click \textquotesingle Subscribe\textquotesingle{} button & Optimal & 8.56 & $-$4.66 & Correct & 2.90 & Optimal Correct \\
\rowcolor{green!8} 2 & Click \textquotesingle Subscribe\textquotesingle{} button (edge) & Optimal & 8.50 & $-$4.62 & Correct & 2.90 & Optimal Correct \\
\midrule
\rowcolor{blue!8} 7 & Click $\vdots$ channel menu & Suboptimal & 7.66 & $-$5.09 & Wrong & 0.80 & Semantic Error \\
\rowcolor{blue!8} 8 & Click OpenAI avatar & Suboptimal & 7.47 & $-$5.09 & Wrong & 0.30 & Semantic Error \\
\midrule
\rowcolor{orange!10} 10 & Click $\times$ clear search & Distractor & 7.38 & $-$3.98 & Wrong & 0.35 & Semantic Error \\
\rowcolor{orange!10} 11 & Click search bar (right) & Distractor & 7.12 & $-$3.92 & Wrong & 0.20 & Semantic Error \\
\rowcolor{orange!10} 12 & Click search bar (left) & Distractor & 6.81 & $-$4.16 & Wrong & 0.40 & Semantic Error \\
\midrule
\rowcolor{gray!8} 28 & Navigate home & Unrelated & 6.78 & $-$4.56 & Wrong & $-$0.90 & Unrelated \\
\rowcolor{gray!8} 34 & Navigate back & Unrelated & 5.91 & $-$4.56 & Wrong & 0.20 & Semantic Error \\
\rowcolor{gray!8} 35 & Open app (empty) & Unrelated & 5.31 & $-$3.81 & Wrong & $-$0.80 & Unrelated \\
\rowcolor{gray!8} 36 & Type empty string & Unrelated & 5.28 & $-$4.84 & Wrong & 0.30 & Semantic Error \\
\bottomrule
\end{tabular}%
}
\end{table}

Three observations stand out. First, \textbf{BBCritic's scores decrease monotonically across the four taxonomy levels} on this page (Optimal $\approx 8.5$, Suboptimal $\approx 7.5$, Distractor $\approx 7.0$--$7.4$, Unrelated $\approx 5.3$--$6.8$), preserving the full semantic hierarchy without any explicit multi-level supervision during training. Second, \textbf{GAIA's binary judgment collapses the entire non-Optimal half of the spectrum into a single ``Wrong'' bucket}: all six Suboptimal/Distractor/Unrelated examples receive the same hard rejection, and the underlying logits offer no usable ordering---the Suboptimal candidates (rows 7--8) actually receive the \textit{lowest} GAIA scores ($-5.09$), worse than several Distractor and Unrelated candidates. This is a direct, per-candidate manifestation of the \textit{Affordance Collapse} described in Sec.~\ref{sec:binary_mismatch} and the distributional entanglement visualized in Figure~\ref{fig:motivation}(c). Third, \textbf{even GPT-5 fails to recover the fine-grained ordering}: it correctly identifies the two Optimal actions but conflates Suboptimal (rows 7--8) and Distractor (rows 10--12) under the same ``Semantic Error'' label, and inconsistently labels Unrelated actions as either ``Unrelated'' or ``Semantic Error'' (e.g., rows 34 and 36). This shows that fine-grained GUI affordance modeling is not solved merely by stronger general capabilities; it requires a critic objective that explicitly preserves continuous ranking structure, which is what BBCritic provides.

\subsection{Implications}

The case study above is a single-page instance of the same phenomenon that Sec.~\ref{motivation} establishes at the dataset level: binary objectives cannot resolve the most ambiguous boundaries (Suboptimal vs.\ Distractor, Distractor vs.\ Unrelated), and zero-shot prompting of a strong VLM is also insufficient. By contrast, BBCritic's contrastive metric-learning formulation produces continuous scores that respect the underlying functional hierarchy, which is precisely the property required for ranking-based test-time scaling.

\section{Broader Impact and Extended Limitations} \label{app:impact}

\subsection{Broader Impact}

This work aims to improve the reliability of autonomous GUI agents through more accurate action evaluation. Positive applications include enhanced accessibility tools for users with disabilities, more reliable automated testing pipelines, and safer human--computer interaction in critical workflows. However, improved GUI automation also carries risks: it could be misused for unauthorized access, social engineering attacks via automated UI interaction, or large-scale web scraping that violates terms of service. We encourage the research community to develop appropriate safeguards alongside capability improvements.

\subsection{Extended Limitations}

Beyond the limitations discussed in the main text, we note several additional considerations:

\textbf{Application scope beyond TTS.} While this work demonstrates BBCritic in the Test-Time Scaling setting, the metric-learning formulation generalizes naturally to other critic-driven scenarios. (i) \textit{Training data curation}: continuous critic scores can rank synthetic trajectories for SFT or preference data filtering. (ii) \textit{Online RL reward modeling}: the critic provides dense per-step signals beyond sparse task-completion rewards. (iii) \textit{Agent self-correction}: critic confidence margins can trigger reflection or replan during execution. Validating these directions requires substantial RL infrastructure (rollout farms, reward shaping pipelines) beyond the scope of this study, and we leave systematic exploration to future work.

\textbf{Platform coverage.} All training data comes from mobile platforms (Android). While zero-shot transfer to Desktop and Web is encouraging (14.2\% improvement on ScreenSpotV2), incorporating multi-platform training data would likely improve grounding accuracy on these domains, particularly for platform-specific UI conventions (e.g., right-click menus on Desktop, hover states on Web).

\textbf{Suboptimal sample scarcity.} The 652 Suboptimal samples (3.6\%) in BBBench reflect both the natural rarity of redundant-but-valid paths on typical GUI pages and the human verification cost of execution-grounded annotation required to identify them reliably. Scaling up rare-category coverage at low cost remains an open community challenge that affects all fine-grained GUI critique benchmarks.

\textbf{VLM rollout dependency.} The diversity and difficulty of Stage 2 hard negatives are shaped by the VLM policy used for rollout: a weaker policy yields easier negatives, while a stronger policy yields negatives closer to the ground truth. Both regimes are usable in practice, and adaptive negative mining strategies that adjust difficulty based on the current critic's competence are a promising direction for further sharpening boundary discrimination.

\textbf{Positional information.} Our semantic embedding compresses fine-grained positional information, which contributes to the slightly smaller Mobile-Icon gain on ScreenSpotV2 where precise grounding is critical. Incorporating explicit spatial features or position-aware contrastive objectives could further improve the trade-off between semantic richness and spatial precision.

\section{Evaluation Details} \label{app:eval}

\subsection{BBBench Evaluation Protocol}

\textbf{NDCG Computation.} We use the exponential gain function $rel(a) = 2^{\ell(a)} - 1$, where $\ell(a) \in \{3, 2, 1, 0\}$ maps to Optimal, Suboptimal, Distractor, and Unrelated respectively. The discount function is $\log_2(r + 1)$ where $r$ is the rank position. We report NDCG@8, NDCG@16, and NDCG@All to evaluate both local retrieval utility and global ranking capability.

\textbf{PPA Computation.} For each pair of adjacent taxonomy tiers $(\mathcal{U}, \mathcal{V})$, we enumerate all $(u, v)$ pairs where $u \in \mathcal{U}$ and $v \in \mathcal{V}$ belong to the same page. PPA is defined as $\mathbb{E}[\mathbb{I}(s_u > s_v) + 0.5 \cdot \mathbb{I}(s_u = s_v)]$. A score of 50\% indicates chance-level discrimination; 100\% means the model perfectly separates adjacent tiers.

\textbf{Scoring Functions.} For binary baselines (e.g., GAIA, GUI-Critic-R1), we use $\sigma(\text{logit}_{\text{Correct}})$ as the critic score, where $\sigma$ is the sigmoid function and $\text{logit}_{\text{Correct}}$ is the logit value of the ``Correct'' token. For BBCritic, we use $\cos(v_i, v_a^i) / \tau$ as defined in the main text (Sec.~\ref{hypos}). For general VLMs evaluated via prompting, we use a 0--10 Likert scale score extracted from the model's text output.

\subsection{TTS Evaluation Setup}

\textbf{Candidate Generation.} For offline TTS (AndroidControl, GUI Odyssey, ScreenSpotV2), the policy model generates $N=8$ candidate actions per step via temperature sampling (temperature 1.0). The critic ranks all candidates and selects the top-1.

\textbf{Margin Metric.} We define the decision margin as $\text{Margin} = \bar{s}_{\text{correct}} - \bar{s}_{\text{incorrect}}$, where $\bar{s}$ denotes the arithmetic mean critic score over correct and incorrect candidates respectively, computed across all evaluation samples. A positive margin indicates that the critic can, on average, distinguish correct from incorrect actions; a larger margin implies more confident decisions and greater robustness to noise. This is the metric reported in Tables~\ref{tab:ablation-stage}, \ref{tab:data_scale}, and~\ref{tab:hard_neg}.

\end{document}